\pdfoutput=1
\documentclass[5p,final]{elsarticle}

\usepackage{elsdouble}
\usepackage{framed,multirow}
\usepackage{hyperref}
\usepackage{lineno}
\modulolinenumbers[1]

\usepackage[small]{caption}
\usepackage{graphicx}
\usepackage{amsmath}
\usepackage{amsthm}
\usepackage{booktabs}
\usepackage{algorithm}
\usepackage{algorithmic}
\usepackage[table,x11names]{xcolor}
\definecolor{LightCyan}{rgb}{0.9,0.96,1}
\usepackage{caption}
\usepackage{subcaption}
\usepackage{amsfonts}
\usepackage{amssymb}
\usepackage{amsmath}
\usepackage{listings}
\usepackage{pifont}
\usepackage{bm}
\usepackage{bbding}
\hypersetup{pdfauthor=author}

\newcommand{\xmark}{\ding{55}}%
\newcommand{\cmark}{\ding{51}}%

\begin{document}

\begin{frontmatter}

\title{Hyperspherically Regularized Networks for Self-Supervision}

%% Group authors per affiliation:
\author[address]{Aiden Durrant}\corref{cor}
\cortext[cor]{Corresponding author}
\ead{a.durrant.20@abdn.ac.uk}

\author[address]{Georgios Leontidis}

\address[address]{Department of Computing Science \& Interdisciplinary Centre for Data and AI, University of Aberdeen, AB24 3UE, Aberdeen, United Kingdom}

\begin{abstract}
Bootstrap Your Own Latent (BYOL) introduced an approach to self-supervised learning avoiding the contrastive paradigm and subsequently removing the computational burden of negative sampling associated with such methods. However, we empirically find that the image representations produced under the BYOL's self-distillation paradigm are poorly distributed in representation space compared to contrastive methods. This work empirically demonstrates that feature diversity enforced by contrastive losses is beneficial to image representation uniformity when employed in BYOL, and as such, provides greater inter-class representation separability. Additionally, we explore and advocate the use of regularization methods, specifically the layer-wise minimization of hyperspherical energy (i.e. maximization of entropy) of network weights to encourage representation uniformity. We show that directly optimizing a measure of uniformity alongside the standard loss, or regularizing the networks of the BYOL architecture to minimize the hyperspherical energy of neurons can produce more uniformly distributed and therefore better performing representations for downstream tasks.
\end{abstract}

\end{frontmatter}

% \linenumbers

\section{Introduction}
Unsupervised visual representational learning methods~\cite{chen2020simple, grill2020bootstrap} have recently demonstrated performance on downstream tasks that continues to narrow the gap to supervised pre-training, excelling specifically in classification and segmentation tasks \cite{caron2020unsupervised, caron2021emerging, zhang2022self}. This success is largely contributed to contrastive methods which aim to minimize the distance of the representations pertaining to two \textit{views} of the same image in representations space (`positive pair'), whilst maximizing the distance of views from different images (`negative pair')~\cite{chopra2005learning}. This ensures that semantically relevant features encoded by representations of positive pairs are similar, whilst negative pairs are dissimilar.  The study of contrastive losses has shown that this repulsion effect between dissimilar views is matching the distribution of features in representational space to a distribution of high entropy \cite{chen2020intriguing}, in other words, encouraging uniformity of representations in space~\cite{wang2020understanding}. This balancing of attraction and repulsion is the mechanism that allows contrastive methods to learn similar semantic features whilst avoiding collapse in representation space. 

More recently, alternative approaches aim to explore self-supervised learning avoiding the inherent computational difficulties imposed by contrastive methods reliance on negative samples~\cite{caron2020unsupervised}. One method in particular, Bootstrap Your Own Latent (BYOL) \cite{grill2020bootstrap} falls under the self-distillation paradigm, where we task an 'online network` to predict the image representations produced by a 'target' network in a Siamese fashion, where each network is given a different view of the same image (visually depicted in Figure \ref{eq:byol}). Yet this network does away with the negative views (views originating from different images), and the subsequent negative term attributed with contrastive losses. The theoretical understanding of how these networks avoid the seemingly inevitable collapsed equilibria, given no explicit mechanism associated with the negative term of contrastive losses, is still to be investigated \cite{richemond2020byol}. 

Intrigued by the property of mode collapse and inspired by \cite{wang2020understanding}, we empirically demonstrate in this work that BYOL fails to distribute its image representations as uniformly in $\ell_2$-normalized unit space (i.e. surface of a unit-hypersphere) compared to its contrastive counterparts. As such, we ask, \textit{can BYOL benefit from mechanisms that introduce feature uniformity found in contrastive methods?} To achieve this, we investigate an alternative to the uniformity constraint posed by \cite{wang2020understanding} and derived from the contrastive loss, aiming to maintain the avoidance of negative sampling advocated in BYOL. We instead propose to utilize minimum hyperspherical energy (MHE) weight regularization \cite{liu2018learning} to enforce neuron (\textit{i.e.} kernel) uniformity whilst being independent and therefore robust to smaller batch size (the desirable property of BYOL). We empirically demonstrate how the use of MHE regularization can increase uniformity of representation through the concept of neuron redundancy reduction in the $\ell_2$-normalized unit space, and in-turn lead to better learned image representations.
\begin{figure}[t]
  \centering
  \begin{minipage}[b]{0.49\textwidth}
  \centering
    \begin{subfigure}[b]{\textwidth}
    \includegraphics[width=0.95\textwidth]{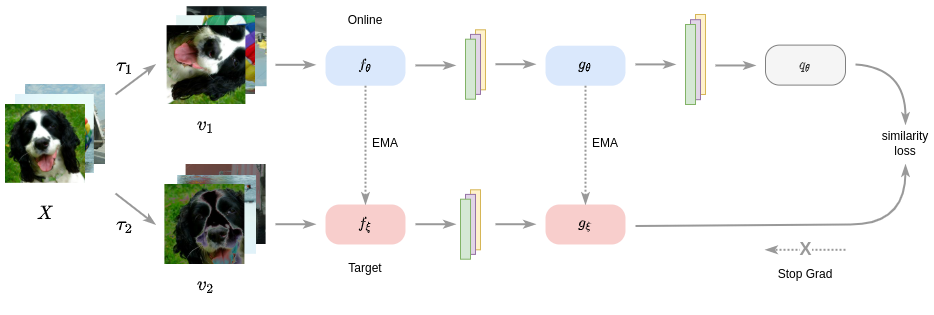}
    % \caption{ }
    \end{subfigure}
  \end{minipage}
\caption{Visual depiction of BYOL architecture \cite{grill2020bootstrap}.}
  \label{fig:byol}
\end{figure}
Our contributions are summarized as follows: i) we empirically show that BYOL distributes its features poorly in representational space compared to contrastive counter parts and that distribution constraints like those in constrastive losses benefit image representations in BYOL; ii) we propose to hyperspherically regularize the network to improve distribution of neurons and subsequently achieve a greater diversity of representations improving image representation separability and performance on downstream tasks; iii) as a consequence hyperspherically regularized BYOL networks maintain the benefits of avoiding contrastive loss negative terms, resulting in reduced performance drops at smaller batch sizes.
\section{Related Work} 
\subsection{Unsupervised Representational Learning}
The recent popularity of discriminative unsupervised representational learning, specifically contrastive methods, has sparked keen interest in the theoretical understanding of their underpinnings, emerging from their performance rivaling that of supervised methodologies \cite{chen2020simple,he2020momentum}. As to why these methods perform so well has only recently begun to be understood, notably \cite{wang2020understanding} prove that optimizing contrastive loss when under a unit $\ell_2$-norm constraint (restricting representational space to a unit hypersphere) is equivalent to optimizing a metric of alignment (distance between positive pairs) and uniformity (all feature vectors should be roughly uniformly distributed on the unit hypersphere). Additionally, \cite{chen2020intriguing} extends this work proposing a generic form of the contrastive loss, also identifying the same relations of uniformity to pairwise potential in a Gaussian kernel, to match representations to a prior distribution (of high entropy).

Lately, alternatives to contrastive methods \cite{caron2020unsupervised} have been proposed alleviating some of the computational drawbacks associated to contrastive losses, primarily the necessity of large numbers of negative pairs generally requiring increased batch sizes \cite{chen2020simple} or memory banks~\cite{he2020momentum}. Bootstrap Your Own Latent (BYOL) avoided the use of negative pairs via an `online' `target' network approach akin to Mean Teachers~\cite{tarvainen2017mean}, where the `online' network and an additional `predictor' network aim to predict the representations of a slowly updated `target' network of the `online' network. However, it is not clear how these networks avoid collapsed representations, it has been hypothesized Batch Normalization (BN) was the critical mechanism preventing collapse in BYOL~\cite{fetterman2020BNBYOL}, yet this hypothesis was refuted, showing batch-independent normalization schemes still achieve comparable performance~\cite{richemond2020byol,chen2021exploring}.

\subsection{Minimal Hyperspherical Energy and Diversity Regularization}
Many unsupervised representational methods learn their representations under the constraint to lie on the surface of a unit-hypershpere via a $\ell_2$-norm constrain leading to desirable traits \cite{xu2018spherical}. As aforementioned \cite{wang2020understanding,caron2020unsupervised} prove that the negative term (repulsion of negative views) in the contrastive loss is equivalent to the minimization of hyperspherical energy of representations. The minimization of hyperspherical energy, the Thompson Problem \cite{thomson1904xxiv}, is a well studied problem in Physics finding the minimal electrostatic potential energy configuration of electrons. Yet this problem has also found place in providing diversity regularization of neurons \cite{liu2018learning,lin2020regularizing},  avoiding undesired representation redundancy. %Many works have also tackled this problem, some promoting large angle/orthogonality \cite{xie2017all} between neurons. Yet
Our work however, investigates whether these regularization methodologies, introducing greater feature diversity and reducing redundancy, can promote more uniformly distributed image representations in BYOL.

\section{Uniform Distribution of Features}
We now define the necessary components of our investigation, specifically, explicit uniformity constraints on the representation space derived from the InfoNCE contrastive loss \cite{oord2018representation} and hyperspherical energy redundancy regularization to enforce representation diversity through neuron uniformity.

\subsection{Contrastive Learning}
We begin by defining the contrastive loss, specifically the InfoNCE loss informally as the softmax cross entropy loss to identify the positive view among the set of unrelated negative views. Formally, we give this in the notation style of \cite{wang2020understanding}, in which the popular case of contrastive loss is considered where an encoder $f:\mathbb{R}^n \rightarrow \mathcal{S}^{m+1}$ is trained and feature vectors are $\ell_2$-normalized onto the unit-hyperspher $\mathcal{S}$ of $m$ dimensions.
\begin{multline}
    \mathcal{L}_{contrastive}(f;\tau,M)  \triangleq \\ 
    \underset{\substack{(x,y)\sim P_{pos}\\ \{x^{-}_{i}\}^{M}_{i=1}\overset{i.i.d}{\sim} P_{data} }}{\mathbb{E}} \left[ - \log \frac{e^{f(x)^\intercal f(y) / \tau}}{e^{f(x)^\intercal f(y) / \tau} + \Sigma_i e^{f(x^-_i)^\intercal f(y) / \tau} } \right]
\label{eq:contrastive}
\end{multline}

where $P_{data}(\cdot)$ is the distribution of data over $\mathbb{R}^n$, $P_{pos}(\cdot, \cdot)$ is the distribution over positive pairs (augmentations $T_1$,$T_2$ of image $X \sim P_{data}$) $\mathbb{R}^n \times \mathbb{R}^n$, $\tau > 0$ is a temperature hyperparameter, and $M \in \mathbb{Z}_+$ a fixed number of negative samples, i.e. $M = 2B-1$ in \cite{chen2020simple} where $B$ is the batch size. Additionally, under the assumption of our $\ell_2$-norm constraint $f(\cdot) \triangleq f(\cdot) / \|f(\cdot)\|_2 $.

\begin{figure}[t]
  \centering
  \begin{minipage}[b]{0.49\textwidth}
  \centering
    \begin{subfigure}[b]{\textwidth}
    \includegraphics[width=\textwidth]{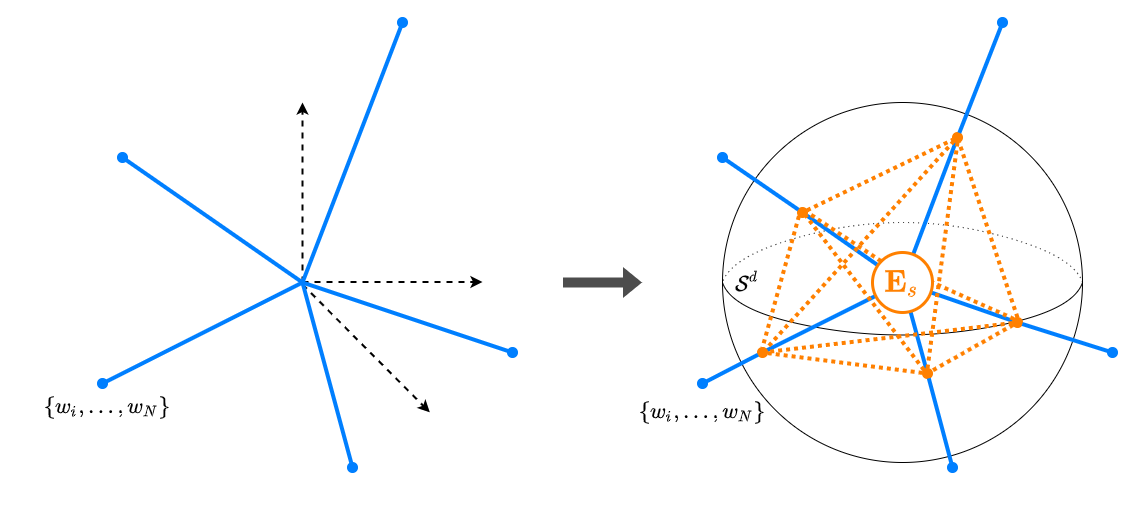}
    % \caption{ }
    \end{subfigure}
  \end{minipage}
\caption{Visual depiction of the regularization of neurons, $\{\bm{w}_1 , \cdots, \bm{w}_N \in \mathbb{R}^{(d+1)}\}$, to minimum hyperspherical energy, $\bm{E}_s$, on the unit hypersphere $\mathcal{S}^d$. \cite{liu2018learning,lin2020regularizing}}
  \label{fig:title_img}
\end{figure}

\subsection{The Link to Uniformity}
It has been shown by the authors of \cite{wang2020understanding} that there is a derivable link to the enforcement of uniformity in contrastive losses. From the loss in Eq.\ref{eq:contrastive}, it is formally shown in \cite{wang2020understanding} that directly optimizing a metric of \textit{alignment} (encourages positive pair representations to be consistent) and \textit{uniformity} (encourages negative pairs to be dissimilar by uniformity distributing representations) is equivalent when $M$ is sufficiently large.
% The \textit{alignment} loss is defined as:
% \begin{equation}
%     \mathcal{L}_{align}(f;\alpha)  \triangleq \underset{(x,y)\sim P_{pos}}{\mathbb{E}} \left[ \| f(x) - f(y) \|_2^\alpha \right] ,  \alpha > 0 , 
% \label{eq:align}
% \end{equation}
% which is simply the expected distance between positive pairs. 

\begin{figure*}[t!]
  \centering
%   \begin{subfigure}[b]{0.6\textwidth}\label{fig:unia}
%   \centering
    
    \begin{minipage}[b]{0.19\textwidth}
    \includegraphics[width=\textwidth]{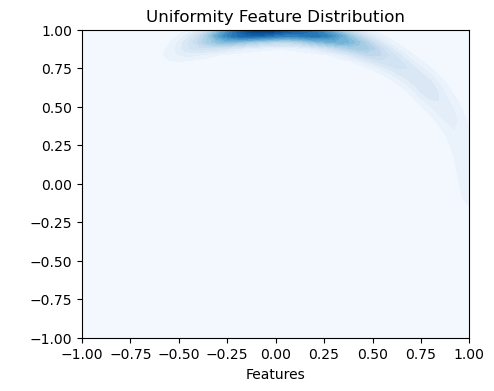}
    \hbox{\hspace{0.15em} \includegraphics[width=0.94\textwidth]{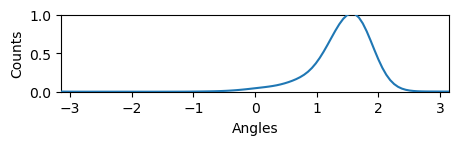}}
    \subcaption{Random Initalization}
    \end{minipage}\hfill
    \begin{minipage}[b]{0.19\textwidth}
    \includegraphics[width=\textwidth]{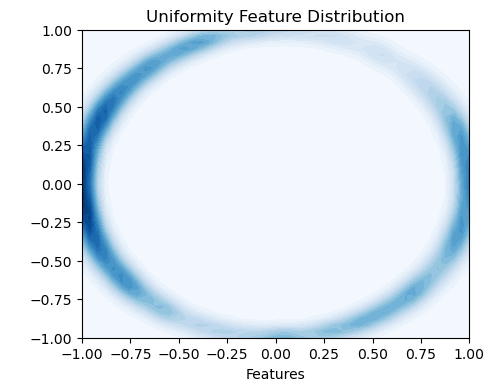}
    \hbox{\hspace{0.15em} \includegraphics[width=0.94\textwidth]{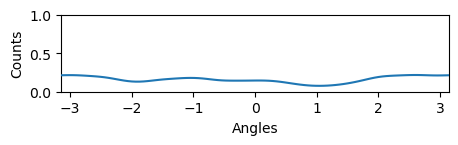}}
    \subcaption{Contrastive Learning}
    \label{fig:uni_b}
    \end{minipage}\hfill
    \begin{minipage}[b]{0.19\textwidth}
    \includegraphics[width=\textwidth]{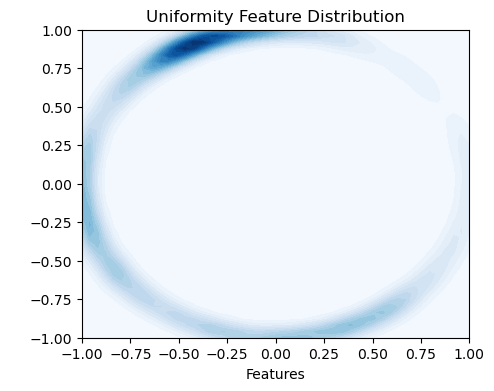}
    \hbox{\hspace{0.15em} \includegraphics[width=0.94\textwidth]{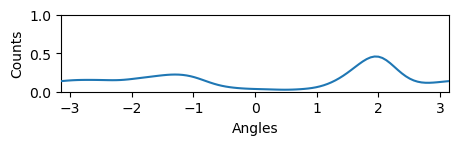}}
    \subcaption{BYOL}
    \label{fig:uni_c}
    \end{minipage}\hfill
    \begin{minipage}[b]{0.19\textwidth}
    \includegraphics[width=\textwidth]{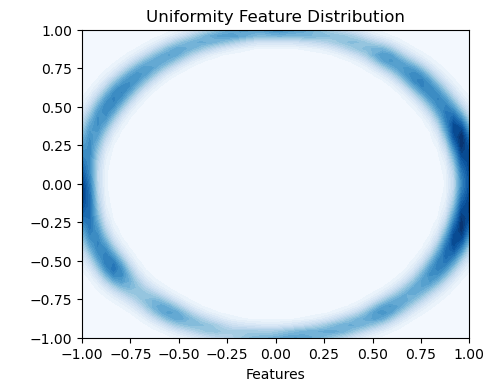}
    \hbox{\hspace{0.15em} \includegraphics[width=0.94\textwidth]{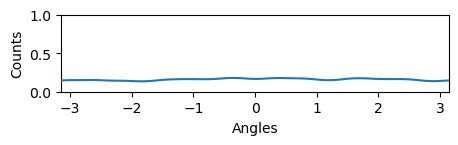}}
    \subcaption{BYOL + Uni}
    \label{fig:uni_d}
    \end{minipage}\hfill
    \begin{minipage}[b]{0.19\textwidth}
    \includegraphics[width=\textwidth]{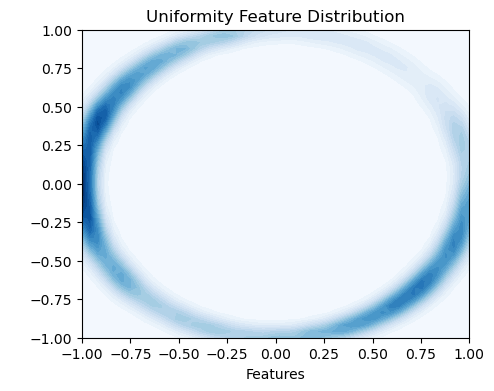}
    \hbox{\hspace{0.15em} \includegraphics[width=0.94\textwidth]{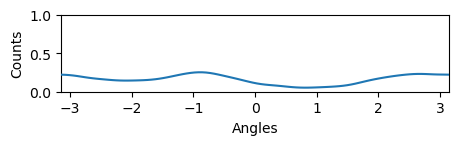}}
    \subcaption{BYOL + MHE Reg }
    \label{fig:uni_e}
    \end{minipage}

  \caption{Learned representations of the CIFAR-10 validation set normalized on the unit hypersphere $\mathcal{S}^1$. The feature distribution is plotted via Gaussian Kernel Density Estimation (KDE) in $\mathbb{R}^2$. The corresponding angles for each $(x,y)$ point in $\mathbb{R}^2$ on the unit hypersphere $\mathcal{S}^1$ is achieved using the von Mises-Fisher KDE~\cite{wang2020understanding}.}
  \label{fig:uni}
\end{figure*}

The \textit{uniformity} loss is given by: 
\begin{equation}
    \mathcal{L}_{uniformity}(f;t)  \triangleq \log \underset{x,y \overset{i.i.d}{\sim} P_{data}}{\mathbb{E}} \left[ e^{-t\|f(x) - f(y)\|^2_2}\right] ,  t > 0 , 
\label{eq:uniformity}
\end{equation}
This derivation, is of our primary interest, where we ponder if the explicit constraint on uniformity can improve representation diversity and subsequently improve performance of BYOL.

\subsection{BYOL and its Uniformity on the Hypersphere}
As previously mentioned, BYOL proposes an alternative to the contrastive paradigm, in which two networks, online ($f_\theta$), and target ($f_\theta$) are each input with a different view of the same image $(x,y)$, with the online network tasked to predict the representations of an identical but temporally aged version of the online network, the target network. The BYOL architecture is depicted in Fig.\ref{fig:byol}, where the prediction is made via a multi-layer perceptron network $q_\theta(f_\theta(\cdot))$ independent of the target network. For specifics regarding BYOL architecture please refer to \ref{sec:implementation} or the original work \cite{grill2020bootstrap}. 

% where the online and target network is extended with a simple multi-layer perceptron projection network, and the prediction is made via another multi-layer perceptron network $q_\theta(f_\theta(\cdot))$ independent of the target network. For specifics regarding BYOL architecture please refer to Section\ref{sec:implementation} or the original work \cite{grill2020bootstrap}. 

The important distinction to the contrastive paradigm regards the loss function Eq.\ref{eq:byol} in which no negative samples are draw (i.e. augmentations/views from different source images). Instead, the loss can be seen as an equivalent to the alignment loss derived from Eq.\ref{eq:contrastive} by \cite{wang2020understanding} (\ref{appendix:alignment}), formally the BYOL loss is given as follows,
\begin{equation}
    \mathcal{L}_{BYOL}(\theta, \xi)  \triangleq \underset{x,y \sim P_{pos}}{\mathbb{E}} \left[ \| \bar{q_\theta}(f_\theta(x)) - \bar{f_\xi}(y)\|^2_2 \right] ,
\label{eq:byol}
\end{equation}

where $\bar{q_\theta}(f_\theta(x)) \triangleq q_\theta(f_\theta(x)) / \|q_\theta(f_\theta(x))\|_2 $ and $\bar{f_\xi}(y) \triangleq f_\xi(y) / \|f_\xi(y)\|_2 $ are the normalization terms projecting the representation onto the unit-hypersphere.

From Eq. \ref{eq:byol} we can observe there exists no term that enforces the separation and therefore diversity of negative views in space. The phenomena associated with the lack of diversity of representations pertaining to different input samples is known as \textit{mode collapse}, where without such term, the network will simple learn trivial and constant representations. The current conjecture as to why BYOL does not exhibit mode collapse lies in the predictor network $q_{\theta}$ \cite{grill2020bootstrap, chen2021exploring} with the derivation and explanation by \cite{grill2020bootstrap} given in \ref{appendix:byol}.

To explore the distribution of representations and subsequently the effect of the predictor $q_{\theta}$ without explicit diversity constraints, we visualize the uniformity of representations produced by the AlexNet~\cite{krizhevsky2017imagenet} encoder when training on the CIFAR-10~\cite{krizhevsky2009learning} dataset. Fig. \ref{fig:uni} depicts the distribution of the validation set features, with Fig.\ref{fig:uni_c} visualizing the distribution under BYOL procedure. We can observe that the distribution of image representations under the BYOL setting, although good and much better than random, is vastly less uniform compared to its contrastive counter-parts.

\subsection{Explicit Uniformity Constraint}\label{section:explicit}
This observation leads to the question: \textit{can BYOL benefit from mechanisms that introduce feature uniformity found in contrastive methods?} More specifically, can the uniformity loss, Eq.\ref{eq:uniformity}, benefit the representations learned under BYOL. This question had been partly explored in~\cite{shi2020run} and indirectly via exploration of negative samples in~\cite{grill2020bootstrap}, yet both of these consider representations produced by the projector and negatives computed from the target network. Given the intuition of BYOL behavior we instead minimize Eq.\ref{eq:uniformity} of the online projections $f_\theta$ only and independent of the predictor $q_\theta$. The intuition behind this procedure is to enforce uniform distribution of features output by the online network, akin to constrastive, whilst maintaining the properties of the predictor to enforce variation via the maximization of information in the uniformly distributed online network. The combined loss is as follows % 
\begin{multline}
   \mathcal{L}_{BYOL + Uni}(\theta, \xi) = \mathcal{L}_{BYOL}(\theta, \xi) + \\ 
   \lambda_{uni} \cdot \mathcal{L}_{uniformity}(f_\theta;t)
\label{eq:byol+uni}
\end{multline}

where $t=2$ and $\lambda_{uni}$ is a hyperparameter controlling the influence of the uniformity metric. 

It can be observed from Fig.\ref{fig:uni_d}, the addition of $\mathcal{L}_{uniformity}$ during BYOL training improves the uniformity of representations produced by the encoder, distributing akin to contrastive (Fig.\ref{fig:uni_b}), confirming expected behavior.

Although the addition of the contrastive uniformity term exhibits greater uniformity of representations and subsequent improvement in performance (Tab. \ref{tab:cifar}), this term relies on the number of negative samples, $M$, being sufficiently large. This is an undesirable computational necessity which BYOL aimed to remove, therefore introducing the contrastive uniformity metric contradicts the advantage of BYOL, the lack of negative samples. Further analysis of the representations captured and robustness to smaller batch size are examined in later sections.

\subsection{MHE Regularization}

It has been shown by~\cite{richemond2020byol,chen2021exploring} that initialization and regularization of weights by batch normalization are fundamental to performant self-supervision. We aim to further extend the power of regularization of self-supervised learning to enforce uniformity of neurons and subsequently the produced representations, whilst avoid the negative sampling constraints imposed by contrastive terms.

We propose the use of hyperspherical regularization~\cite{liu2018learning} alongside batch normalization to explicitly regularize the network to reduce hyperspherical energy of neurons (depicted in Fig.\ref{fig:title_img}) to further improve the diversity of weights in the network and consequently representation uniformity. Fundamentally, such methods aim to reduce undesired representation redundancy occurring through non-uniform distribution of neurons. This choice is particularly motivated the findings in \cite{liu2018learning}, where scenarios of class imbalance where unrepresented classes were shown to be well separated as a result of more uniformly distributed classification neurons. Additionally, \cite{liu2018learning} argues that the power of neural representations can be characterized by the hyperspherical energy of its neurons (\textit{i.e.} kernels), and as such a minimal hyperspherical energy configurations can induce better diversity and improve representation separability. The hyperspherical energy for $N$ neurons, in $\mathbb{R}^{(d+1)}$, $\bm{W}_{N}=\{\bm{w}_1 , \cdots, \bm{w}_N \in \mathbb{R}^{(d+1)}\}$ is defined as:
\begin{multline}
   \bm{E}_{s} = \bm{E}_{s,d}(\hat{\bm{w}}_i |^N_{i=1} ) = \sum\limits_{i=1}^N \sum\limits_{j=1, j \neq i}^N r_s(\| \hat{\bm{w}}_i - \hat{\bm{w}}_j\|) \\
   = \begin{cases}
    \sum_{i\neq j} \| \hat{\bm{w}}_i - \hat{\bm{w}}_j\|^{-s}, & s > 0\\
    \sum_{i\neq j} \log (\| \hat{\bm{w}}_i - \hat{\bm{w}}_j\|^{-1}),               & s=0
    \end{cases}
\label{eq:min}
\end{multline}
where $\hat{\bm{w}}_i=\frac{\bm{w}_i}{\|\bm{w}_i\|}$ is the $i$-th neuron weight projected onto $\mathcal{S}^d$, and $r_s(\cdot)$ is a decreasing real valued function, which is chosen to be the Riesz s-kernel, $r_s(z^{-s}), s > 0$ \cite{liu2018learning}. We therefore aim to minimize the energy $\bm{E}_s$ in Eq.\ref{eq:min} by manipulating the orientation of the neurons $\bm{W}_N$ to solve $\min_{\bm{W}_N}\bm{E}_s$, $s\geq 0$. When $s=0$, the logarithmic energy minimization problem is undertaken, essentially maximizing the product of Euclidean distance, where in our case this is the angle between neurons.
\begin{equation}
   \arg \underset{\bm{W_N}}{\min}\bm{E_0} = \arg \underset{\bm{W_N}}{\min}\exp (\bm{E_0}) = \arg \underset{\bm{W_N}}{\max} \underset{i \neq j}{\prod} \| \hat{\bm{w}}_i - \hat{\bm{w}}_j \|
\label{eq:max}
\end{equation}
As an explicit regularization method, we optimize for the joint objective function:
\begin{equation}
   \mathcal{L} = \mathcal{L}_{BYOL}(\theta, \epsilon) + \lambda_{mhe} \cdot \sum\limits_{j=1}^{L_{\theta}} \frac{1}{N_j (N_j -1)} \{\bm{E}_s\}_j
\label{eq:mhe_reg}
\end{equation}
where $\lambda_{mhe}$ is a hyperparameter to control the weighting of our regularization, $L_{\theta}$ the number of layers in the online network $f_{\theta}$ and/or predictor $q_{\theta}$, and $N_j$ is the number of neurons in layer $j$. A further variant has also been considered in this work simply extending the hyperspherical energy based on Euclidean distance in Eq.\ref{eq:min} to consider geodesic distance on a unit hypersphere. We define this extension in \ref{appendix:algorithm} For more details and all proofs we refer to \cite{liu2018learning}.

\begin{figure}[h]
  \centering
    \centering
    \begin{minipage}[b]{0.235\textwidth}
    \includegraphics[width=\textwidth]{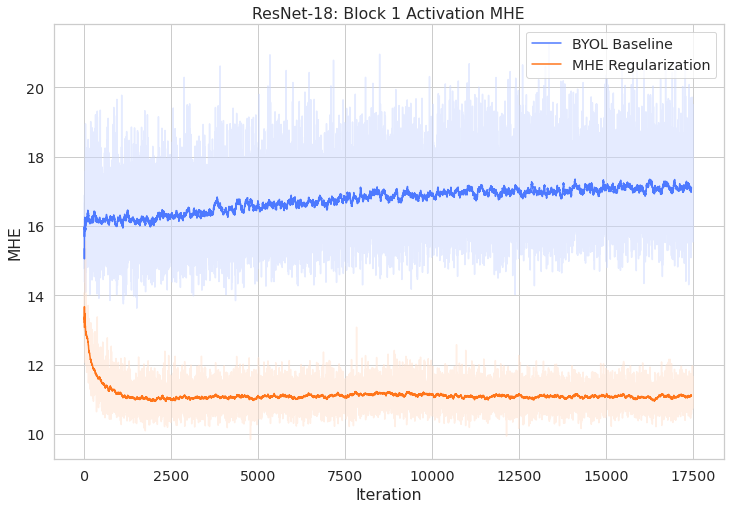}
    \includegraphics[width=\textwidth]{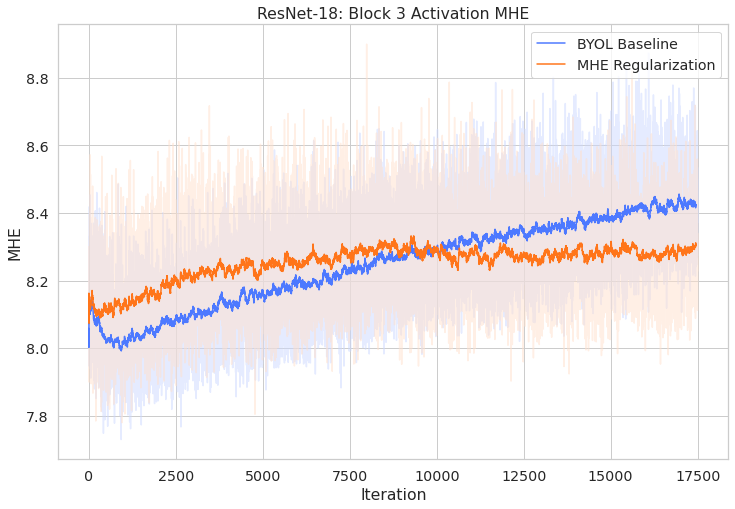}
    \end{minipage}\hfill
    \begin{minipage}[b]{0.235\textwidth}
    \includegraphics[width=\textwidth]{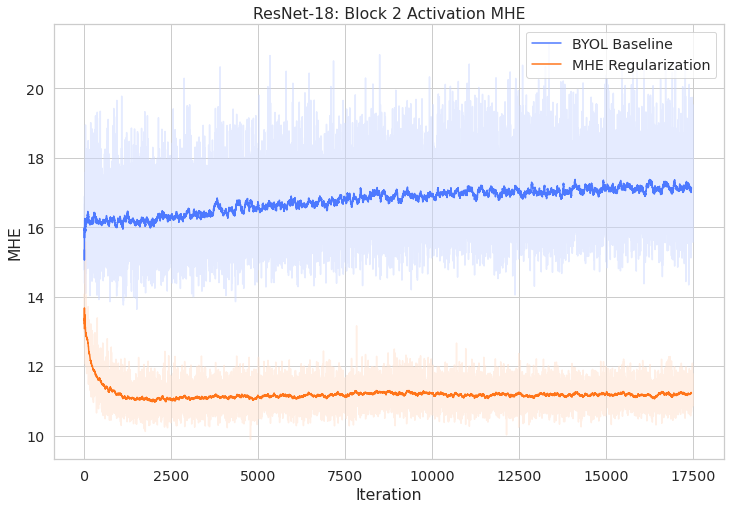}
    \includegraphics[width=\textwidth]{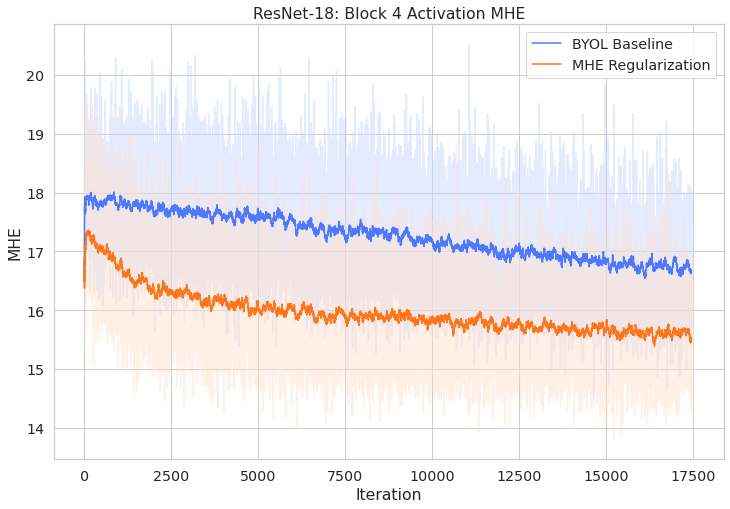}
    \end{minipage}
\small
  \caption{Hyperspherical energy vs. iteration during training for intermediate representations of the ResNet-18 encoder. We compute the MHE on the output of each ResNet block \cite{he2016deep}.}
  \label{fig:layer_uniformity_mhe}
\end{figure}
\subsubsection{Representation Uniformity Analysis}
\begin{figure}[t!]
  \centering
  \vspace{1em}
    \centering
    \begin{subfigure}[b]{0.235\textwidth}
    \includegraphics[width=\textwidth]{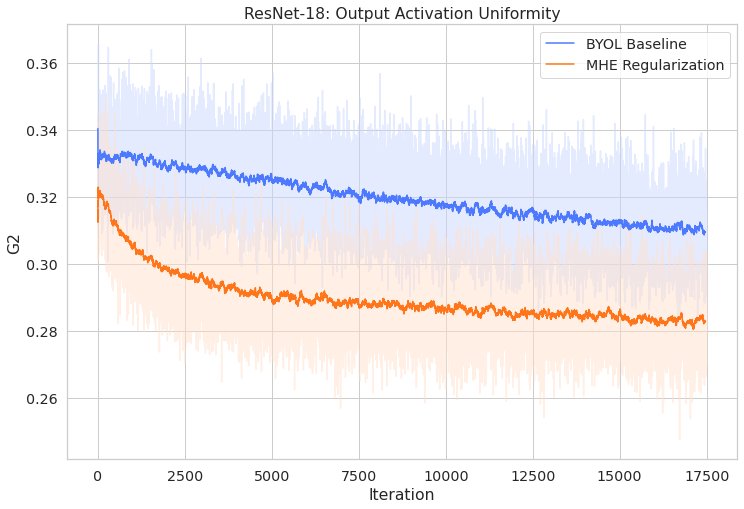}
    \caption{G2}
    \end{subfigure}
    \begin{subfigure}[b]{0.235\textwidth}\label{fig:mhe_opt}
    \includegraphics[width=\textwidth]{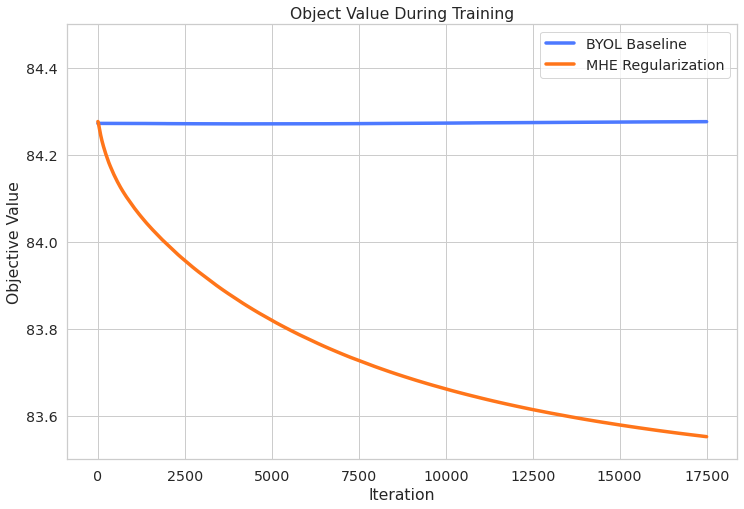}
    \caption{MHE Optimization Value}
    \end{subfigure}
    \small
  \caption{Dynamics during training of the STL-10 dataset. (a) Uniformity measure $G2$ vs. iteration. (b) MHE regularization objective value vs. iteration.}
  \label{fig:g2}
\end{figure}

Demonstrated in Fig.\ref{fig:uni} is the distribution of image representations under MHE regularization following the same visualization methodology presented in \ref{section:explicit}, we empirically confirm our hypothesis that improving the diversity of weights within the network subsequently results in more diversely distributed representations. Fig.\ref{fig:uni_e} show a significant improvement in representation uniformity compared to the baseline in Fig.\ref{fig:uni_c}.
To further confirm these findings, we plot in Fig.\ref{fig:layer_uniformity_mhe} the hyperspherical energy of intermediate layer representations of a ResNet-18 encoder during training on the CIFAR-10 dataset between standard BYOL and BYOL with MHE regularization applied. We empirically show that regularizing the neurons via MHE maintains lower hyperspherical energy on its activation/representations throughout the whole network, compared to BYOL baseline. It is motivating to note that the final output layer representations demonstrate immediately lower hyperspherical energy, and increased uniformity by a measure of $G2$ (\ref{appendix:g2}), Fig.\ref{fig:g2}, throughout training by a considerable margin, an important factor for learning good representations applied to downstream tasks. The empirical finding in Fig.\ref{fig:layer_uniformity_mhe} support our hypothesis and rationale that increasing diversity of weights leads to representations that are in-turn more uniformly distributed. We report performance benchmarks in \ref{sec:linear_eval}, and ablations in. \ref{sec:ablation}.%
\subsection{Implementation Details}\label{sec:implementation}
The implementation follows the procedure presented in \cite{grill2020bootstrap} with exception to the addition of the regularization loss terms. As to correspond with the BYOL procedure, we employ the same image augmentations as described in ~\cite{chen2020simple,grill2020bootstrap}. Similarly, our experimentation primarily focuses on the use of two different convolutional residual network~\cite{he2016deep} configurations for our encoders $f$, ResNet-18 and ResNet-50. Following the procedure described in BYOL~\cite{grill2020bootstrap}, the networks replace the standard linear output layer with a Multi-Layer Perceptron (MLP) $g$, projecting the output of the final average pooling layer to a smaller space. The MLP is a two layer linear network the first outputting in 4096 dimensions, followed by a second outputting to 256 dimensions. The first layer only is followed by batch normalization and Rectified Linear Unit (ReLU) non-linearity. Specifics regarding full augmentation details and optimization settings are given in full in \ref{appendix:implementation details}.
\begin{table}[t]
    \small
    \centering
    \caption{\textbf{ImageNet Linear Classification:} encoder trained for 1000 epochs. We report top-1 accuracy (\%) and $k$-NN accuracy. *=Reproduction, RN50=ResNet-50}
    \begin{tabular}{lcc|cc}
        Method  & Arch. & Batch Size & top-1 & \textit{k}-NN \\
        \midrule
        Supervised  & RN50 & - & 79.3 & 79.3 \\
        SimCLR \cite{chen2020simple} & RN50 & 4096 & 69.1 & 60.7 \\
        MoCov2 \cite{chen2020improved}     & RN50 & 4096 & 71.1 & 61.9 \\
        InfoMin \cite{tian2020makes}    & RN50 & 4096 & 73.0 & 65.3 \\
        BarlowT \cite{zbontar2021barlow}    & RN50 & 4096 & 73.2 & 66.0 \\
        OBoW \cite{gidaris2020online}       & RN50 & 4096 & 73.8 & 61.9 \\
        % DCv2 \cite{caron2020unsupervised}      & RN50 & 4096 & 75.2 & 67.1 \\
        % SwAV \cite{caron2020unsupervised}       & RN50 & 4096 & 75.3 & 65.7 \\
        % DINO \cite{caron2021emerging}       & RN50 & 4096 & 75.3 & 67.5 \\
        SimSiam \cite{chen2021exploring}   & RN50 & 256 & 71.3 & -- \\
        \midrule
        BYOL \cite{grill2020bootstrap}      & RN50 & 4096 & 74.3 & 64.8 \\
        BYOL* \cite{grill2020bootstrap}      & RN50 & 4096 & 74.1 & 63.7 \\
        \cellcolor{LightCyan}BYOL-MHE & \cellcolor{LightCyan}RN50 & \cellcolor{LightCyan}4096 & \cellcolor{LightCyan}\textbf{74.4} & \cellcolor{LightCyan}\textbf{64.9} \\
        % \cellcolor{LightCyan} $\text{MoP}_{BYOL}$ & \cellcolor{LightCyan} RN50 & \cellcolor{LightCyan} \textbf{256} & \cellcolor{LightCyan} 00.00 & \cellcolor{LightCyan} 00.00
    \end{tabular}
    \label{tab:imgnet1000}
\end{table}
\begin{table}[t]
\small
\caption{\textbf{ImageNet Linear Classification:} ResNet-50 encoder trained for 300 epochs. We report top-1 and top-5 Accuracy (\%). *=Reproduction}
\centering
\small
\begin{tabular}{lrr}
\toprule
Method  & Top-1 (\%) & Top-5 (\%) \\
\midrule
SimCLR~\cite{chen2020simple}      & 67.9  & 88.5      \\
BYOL~\cite{grill2020bootstrap}        & 72.5  & 90.8      \\
BYOL*        & 71.9  & 89.2      \\
\cellcolor{LightCyan}BYOL-MHE & \cellcolor{LightCyan} 72.4  & \cellcolor{LightCyan} 89.9 \\
\bottomrule
\end{tabular}
\label{tab:imagenet}
\end{table}%
\section{Linear Evaluation}\label{sec:linear_eval}
To evaluate the quality of representations learned during self-supervised training we employ the standard linear evaluation protocol described in \cite{chen2020simple,grill2020bootstrap}. For context, a linear classifier is trained taking as input the representations produced by the encoder $f_\theta$ which is pre-trained in a self-supervised manner and then frozen as to not train in a supervised manner. Tab.\ref{tab:imgnet1000} reports the top-1 accuracy in \% for the ImageNet ILSVRC-2012 test set trained with a standard ResNet-50 for 1000 epochs, we show our reproduction of methods alongside our MHE regularized variant. We empirically set $\lambda_{uni}=0.125$, and $\lambda_{mhe}=1.$, choosing the angular variant of MHE regularization with power $s=2$ applying to all weights in the encoder, projector and predictor, \ref{appendix:lambda}.

\begin{figure}[t]
  \centering
%   \begin{subfigure}[b]{0.6\textwidth}\label{fig:unia}
%   \centering
    \includegraphics[width=0.30\textwidth]{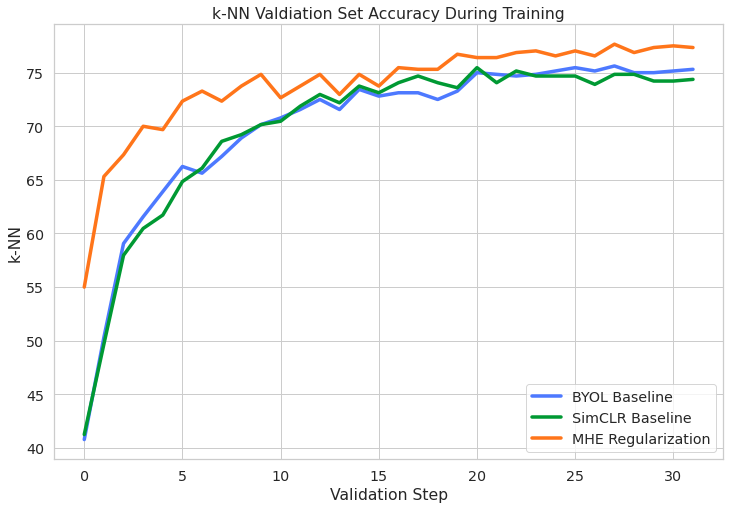}
    \small
  \caption{$k$-Nearest Neighbor accuracy on the STL-10 validation set during training.}
  \label{fig:knn}
\end{figure}

We report \textbf{74.4} top-1 accuracy with the inclusion of MHE regularization, a 0.32\% improvement over the standard BYOL baseline, a significant improvement in performance which matches the jump made between other competing methodologies. Additionally, the improvement is maintained at lower epoch counts, and smaller batch sizes, with Tab.\ref{tab:imagenet} demonstrating 0.5\% improvement at 300 epochs and 1024 batch size. Robustness to hyperparameters is further explore in \ref{sec:ablation}, yet these benchmark results are a substantial and clear improvement, further validating the capability of our regularization to provide better representations with minimal overhead. In addition to top-1, we report the weighted nearest neighbor classifier, $k$-NN, as a measure of representation semantic alignment and separability with very little variation due to hyperparameter values. Tab.\ref{tab:imgnet1000} reports a large improvement over baseline by 1.0\%, and similarly Fig.\ref{fig:knn} mirrors such findings when training on STL-10. 

In the case of CIFAR-10, CIFAR-100 and STL-10 datasets we see a more substantial improvement, reporting a top-1 accuracy of 94.78\%, and 72.56\% for CIFAR-10 and CIFAR-100 respectively. This 0.4\% improvement in CIFAR-10 is comparable to the improvements found between SimCLR and BYOL, a substantial move towards the supervised baseline of 95.1\% reported in \cite{chen2020simple}. For the explicit uniformity constraint, BYOL + $\mathcal{L}_{uniformity}$, we see on average 0.1\% improvement from the MHE regularized variant. This improvement is expected, given the more explicit nature of the uniformity constraint on directly optimizing the representations rather than the implicit MHE regularization, and the observed near uniform distribution depicted in Fig.\ref{fig:uni_d}. Interestingly, we find the MHE regularization to be the best performing setting on the STL-10 dataset, with a 1.1\% improvement over baseline, and 0.8\% improvement over the explicit uniformity constraint. We can conjecture that the large and diverse nature of the semantic classes in STL-10 unlabeled set benefit more from the unique representation neuron effect that enables unrepresented concepts/classes to be uniquely and evenly assigned \cite{liu2018learning}.
\begin{table}[t]
\small
\caption{Top-1 (\%) Accuracies of Linear Evaluation on CIFAR10 and CIFAR100 datasets with ResNet-50 Encoder trained for 1000 epochs. $* =$ reproduction }
\centering
\small
\begin{tabular}{lrrr}
\toprule
Method  & CIFAR 10 & CIFAR 100 & STL-10 \\
\midrule
% Supervised */(repro)  & 95.1/95.62  \\
SimCLR*   & 93.81 & 70.98 & 82.40 \\
BYOL*    & 94.46 & 72.10 & 82.81 \\
\cellcolor{LightCyan}BYOL + $\mathcal{L}_{Uni}$      & \cellcolor{LightCyan} \textbf{94.84} & \cellcolor{LightCyan}\textbf{72.62} & \cellcolor{LightCyan}83.19 \\
\cellcolor{LightCyan}BYOL-MHE        & \cellcolor{LightCyan}94.78 & \cellcolor{LightCyan}72.56 & \cellcolor{LightCyan}\textbf{83.96} \\
\bottomrule
\end{tabular}
\label{tab:cifar}
\end{table}
\section{Ablation and Sensitivity Analysis}\label{sec:ablation}
We analyze the behavior of our BYOL constraints exploring the impact of hyperparameter and network configurations. We follow the procedure described in \ref{sec:implementation} and \ref{sec:implementation}, training a ResNet-18 encoder for 300 epochs.%
\begin{figure}[t]
  \centering
    \begin{minipage}[b]{0.49\textwidth}
    \centering
    \begin{subfigure}[b]{0.35\textwidth}
    \includegraphics[width=\textwidth]{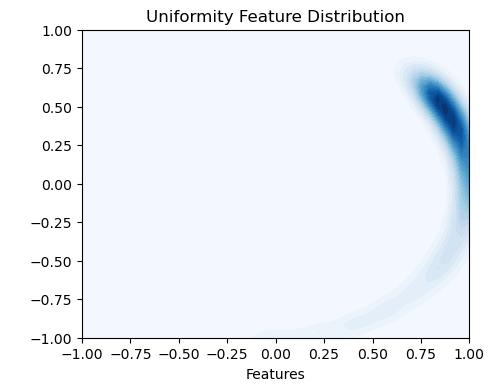}
    \caption{No BN + No MHE}
    \end{subfigure}
    \begin{subfigure}[b]{0.35\textwidth}
    \includegraphics[width=\textwidth]{BYOL_Uniformity_f.png}
    \caption{BN + No MHE}
    \end{subfigure}
    \end{minipage}
    \begin{minipage}[b]{0.49\textwidth}
    \centering
    \begin{subfigure}[b]{0.35\textwidth}
    \includegraphics[width=\textwidth]{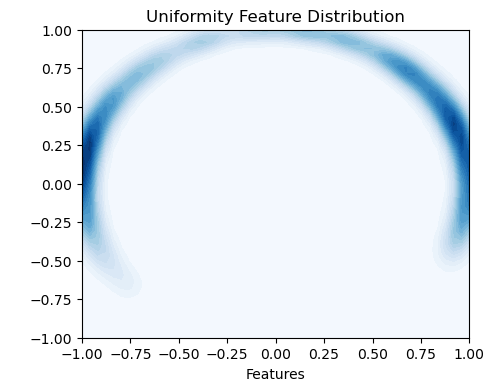}
    \caption{No BN + MHE}
    \label{fig:pca_c}
    \end{subfigure}
    \begin{subfigure}[b]{0.35\textwidth}
    \includegraphics[width=\textwidth]{BYOL-mhe-a2_Uniformity_f.png}
    \caption{BN + MHE}
    \end{subfigure}
    \end{minipage}

  \caption{Uniformity of representations on $\mathcal{S}^1$ under different regularization configurations plotted with Gaussian KDE.}
  \label{fig:pca}
\end{figure}
\subsection{Batch Size}\label{sec:batch}
\begin{figure}[t]
  \centering
%   \begin{subfigure}[b]{0.6\textwidth}\label{fig:unia}
%   \centering
    \includegraphics[width=0.3\textwidth]{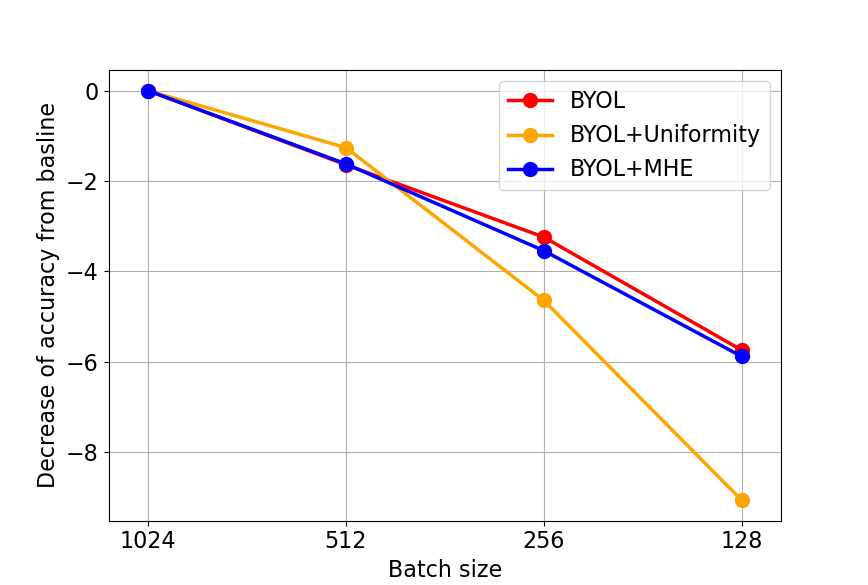}
\small
  \caption{Reduction in CIFAR-10 linear evaluation top-1 \% when decreasing batch size for the proposed variants and repro of BYOL.}
  \label{fig:batch}
\end{figure}

One primary advantage BYOL introduced is the robustness to smaller batch sizes, this emerges from the avoidance of negative pairs sampled from within the batch in end-to-end contrastive models. Therefore, with our addition of $\mathcal{L}_{uniformity}$ (Eq.\ref{eq:uniformity}) being derived from Eq.\ref{eq:contrastive}, we expect robustness to batch size to degrade. We test the performance under different batch size averaging gradients over $N$ consecutive steps before updating the network parameters, where $N$ is the factor of batch size reduction from the baseline~\cite{grill2020bootstrap}. Fig.\ref{fig:batch} shows that the introduction of the explicit uniformity loss reduces robustness to batch size as expected. We see from a baseline of 91.48\%, a -9.06\% drop with $\mathcal{L}_{uniformity}(f_\theta)$ compared to BYOL's -5.74\%. This expected result confirms our reasoning to find alternative mechanisms to enforce uniformity of image representations. For MHE regularization, we observe little deviation of performance compared to standard BYOL given the regularization's independence on batch size.
%
%\footnote{All models employ batch normalization, we can partially attribute drop in performance due to computations of batch statistics.}.
%
%
\begin{table}[h]
\small
\caption{Linear Evaluation on CIFAR10 under different regularization configurations, all networks $f_{\theta},g_{\theta},q_{\theta},$ are regularized when selected.}
\centering
\small
\begin{tabular}{lrr}
\toprule
BN  & MHE & Accuracy (\%)\\
\midrule
\xmark & \xmark & 28.76 \\
\cmark  & \xmark & 90.74\\
\xmark      & \cmark &\cellcolor{LightCyan} 45.72  \\
\cmark  & \cmark & \cellcolor{LightCyan}91.22\\
\bottomrule
\end{tabular}
\label{tab:bn}
\end{table}
\subsection{MHE Regularization and Batch Normalization}\label{sec:reg}
Following our intuition and empirical findings that MHE regularization encourages representation uniformity, we further investigate the effect of regularization components. We first explore the uniformity of CIFAR-10 validation set representations as done in Fig.\ref{eq:uniformity}. We can see the representations in $\mathcal{S}^1$ plotted in Fig.\ref{fig:pca} and their corresponding linear evaluation results in Tab.\ref{tab:bn}. The results empirically show how without batch normalization the network fails to learn whilst poorly distributing representations in space, resulting in collapsed representations coinciding with~\cite{richemond2020byol}. Confirming our previous empirical results, BYOL with MHE regularized alone produces image representations that are distributed far more uniform than batch norm, Fig.\ref{fig:pca_c}. However, the linear evaluation performance suffers compared to batch normalization, although our regularization has a similar effect to batch norm in avoiding collapse of representations, albeit with less impact. We conclude from these finding that regularization is a key component to avoid mode collapse in self-distillation methods, where batch normalization is not a fundamental necessity rather the diversity of neurons and reduction in redundancy provided by MHE is enough to encourage variation in learned representations. This is a promising finding which warrants further investigation in future work.

\subsection{MHE Regularization Parameterization}
To investigate how varying hyperparameters for the MHE regularization affects performance, we report results for network configurations in Tab.\ref{tab:sections}. Additionally, the weight of the regularization $\lambda_{mhe}$ and powers $s$ are given in \ref{appendix:hyperparams}. 

We report in Tab.\ref{tab:sections} the linear evaluation performance under varying configurations of MHE regularization to individual sub-networks. We show that across all configurations we see an increase in performance, showing that the improved weight diversity and subsequent representation diversity improves the quality of representations learned. Additionally, referring to our previous notion that it is not preferable to directly enforce uniformity at the predictor of the BYOL architecture based on the intuition of BYOL's behavior~\cite{grill2020bootstrap}, we do not see any degradation in performance when MHE is applied at the predictor level. We conjecture that the improved diversity of features helps assist the online network in capturing more varied representations.

\begin{table*}[t!]
\small
\caption{Linear Evaluation on CIFAR10 given different network configurations of MHE regularization, `\cmark' denotes that MHE regularization has been applied to that sub-network. The encoder is ResNet-18 trained for 300 epochs. \Cross $=$ BYOL (repro)}
\centering
\small
% \begin{tabular}{l||c|c|c|c|c|c|c|c}
\begin{tabular}{l||*{7}{>{\centering\arraybackslash}p{0.05\textwidth}|}p{0.05\textwidth}}
\toprule
\multicolumn{1}{l}{Layer} & \multicolumn{8}{c}{} \\ \midrule
Encoder $f_{\theta,\xi}$ & \multicolumn{4}{c|}{-} & \multicolumn{4}{c}{\cmark} \\
Projector $g_{\theta,\xi}$ & \multicolumn{2}{c|}{-} & \multicolumn{2}{c|}{\cmark}  & \multicolumn{2}{c|}{-} & \multicolumn{2}{c}{\cmark} \\ 
Predictor $q_{\theta}$ & - & \cmark & - & \cmark & - & \cmark & - & \cmark \\ \midrule
\multicolumn{1}{l}{MHE}  & \multicolumn{8}{c}{Accuracy (\%)} \\ \midrule
MHE (a2) & $90.74^{\text{\Cross}}$ & \cellcolor{LightCyan} \textbf{91.64} &  \cellcolor{LightCyan} 91.36 & \cellcolor{LightCyan} 91.46 & \cellcolor{LightCyan} 91.38 & \cellcolor{LightCyan} 91.10 & \cellcolor{LightCyan} 90.96  & \cellcolor{LightCyan} 91.22\\ \midrule
BYOL + $\mathcal{L}_{Uni}$ & 91.48 &   &   &   &   &   &   &  \\
\bottomrule
\end{tabular}
\label{tab:sections}
\end{table*}

\section{Conclusion}
We empirically show that uniformity constrains like those in contrastive losses can be beneficial in BYOL and self-distillation methods in general where negative samples are negated. To maintain the computation benefits proposed by BYOL we investigate the use of regularization methods that minimize the hyperspherical energy between network neurons. We show that this type of redundancy regularization implicitly improves distribution uniformity representations learned by the encoder, leading to improved results in all experimentation over the baseline whilst remaining robust to changes in batch size, with minimal additional computational. Empirical exploration demonstrates the degree in which MHE regularization impacts the uniformity of representations during training throughout the encoder network, validating our intuition that more diverse neurons result in more diverse representations. 

Performance gains from our regularization are significant given no architectural change, nor augmentation change, common in alternative approaches. We believe further performance improvements can be made with tuning of hyperparameters. Yet how the avoidance of fully collapsed equilibria in the presence of MHE regularization identified in this work is still yet to be understood, as is how the maximization of kernel diversity improves activation diversity. However, from this work we have identified the importance of regularization in self-supervision and its effect on learned image representations in space. 
\subsection{Future Work}
This works empirically identifies unexpected training behavior of the self-supervised, self-distilled method BYOL, and as such expanding this exploration to alternative methods is a natural continuation. In addition, the further analysis of regularization in self-supervision as a whole is an importance next step to understand the training dynamics. Furthermore, the identified phenomena shows such regularization impacting uniformity may be enough to solely avoid mode collapse currently prevented by the predictor network\cite{chen2021exploring}, establishing the hypothesis for future investigations.
\section*{Declaration of Competing Interest}
The authors declare that they have no known competing financial interests or personal relationships that could have appeared to influence the work reported in this paper.
\section*{Acknowledgments}
This work used the Cirrus UK National Tier-2 HPC Service at EPCC (http://www.cirrus.ac.uk). Access granted through the project: ec173 - Next gen self-supervised learning systems for vision tasks.

\bibliography{ref}

\pagebreak
\appendix 

\section{Method}

\subsection{Alignment Term}\label{appendix:alignment}
The contrastive loss Eq.\ref{eq:contrastive} is presented by \cite{wang2020understanding} to be equal to optimizing two metrics, the first being uniformity Eq.\ref{eq:uniformity} and the second, alignment which is given as follows:
\begin{equation}
    \mathcal{L}_{align}(f;\alpha)  \triangleq \underset{(x,y)\sim P_{pos}}{\mathbb{E}} \left[ \| f(x) - f(y) \|_2^\alpha \right] ,  \alpha > 0 , 
\label{eq:align}
\end{equation}

\textit{Alignment} encourages positive pair representations to be consistent as to bring together in unit-space the representations pertaining to the same source image. This essentially clusters similar semantic representations together while uniformity distributes the representations in space as to avoid trivial and consistent representations.

\subsection{G2 Measure of Hyperspherical Energy}\label{appendix:g2}
$G2$ refers to the Gaussian potential kernel (radial basis function kernel) with the formulation:
\begin{equation}
  G_t(x,y) = e^{-t\|f(x) - f(y)\|^2_2}, t>0
\end{equation}
This defines the uniformity loss in Eq.\ref{eq:uniformity}, where we simply take the logarithm of the average.

\subsection{BYOL Predictor Network Behavior}\label{appendix:byol}
The current conjecture as to the behavior of BYOL is presented in~\cite{grill2020bootstrap} hypothesizing that BYOL works as a form of dynamic system where the target parameters $\xi$ updates are not in the direction of $\nabla_\xi \mathcal{L}_{BYOL}(\theta, \xi)$. Generally stating, there is no loss $\mathcal{L}(\theta, \xi)$ where BYOL's dynamics is a gradient decent on $\mathcal{L}$ jointly over $\theta, \xi$ (minimization of both target and online). For full details we refer to the original hypothesis \cite{grill2020bootstrap}. More recently work by \cite{chen2021exploring} determined the stop-gradient operation is fundamental to the avoidance of mode collapse, where not present degenerative solutions are found.

\subsection{Angular MHE Regularization}\label{appendix:algorithm}

The hyperspherical energy defined in Eq.\ref{eq:min} is based on Euclidean distance on a hypersphere, however, \cite{liu2018learning} propose an alternative to Euclidean distance measure. The proposed is a simple extension defining the hyperspherical energy based on geodesic distance, replacing $\|\hat{\bm{w}}_i -  \hat{\bm{w}}_j \|$ with $\arccos(\hat{\bm{w}}_i^{\intercal} \hat{\bm{w}}_j)$. The main difference lies in optimization dynamics reported in \cite{liu2018learning}. The extension known as angular MHE is defined as:
\begin{multline}
  \bm{E}_{s}^a = \bm{E}_{s,d}^a(\hat{\bm{w}}_i |^N_{i=1} ) = \sum\limits_{i=1}^N \sum\limits_{j=1, j \neq i}^N r_s(\arccos(\hat{\bm{w}}_i^{\intercal} \hat{\bm{w}}_j)) \\
  = \begin{cases}
    \sum_{i\neq j} \arccos(\hat{\bm{w}}_i^{\intercal} \hat{\bm{w}}_j)^{-s}, & s > 0\\
    \sum_{i\neq j} \log (\arccos(\hat{\bm{w}}_i^{\intercal} \hat{\bm{w}}_j)^{-1}),               & s=0
    \end{cases}
\label{eq:min_angular}
\end{multline}

\section{Implementation Details}\label{appendix:implementation details}
We give full details pertaining to architecture, optimization and data augmentations for all experimental settings.

\subsection{Architecture}
Our experimentation primarily focuses on the use of two different convolutional residual network~\cite{he2016deep} configurations for our encoders $f_\theta$ and $f_\xi$, ResNet-18 and ResNet-50 each with 18 and 50 layers respectively. For the CIFAR datasets we adjust the first layers (\textit{i.e.} `stem') of the ResNet architecture, reducing kernel size of the first convolutional layer to 3 from 7, kernel stride from 2 to 1, and remove the max-pooling operation to accommodate the reduced image size. When reporting on MHE regularization, unless stated otherwise, regularization is applied to the all linear layers in $g_{\theta}$ and $q_{\theta}$.
\subsection{Optimization}
For all unsupervised training we use the LARS optimizer excluding both batch normalization and bias parameters, with cosine learning rate decay, and the learning rate linearly scaled with the batch size $(LR = LR_{base} \times \text{BatchSize}/256)$. Like similar works, we employ 10 linear warm-up epochs to assist large batch-size training. The EMA parameter $\tau$, is increased during training to 1 from $\tau_{base}$ with $\tau \triangleq 1 - (1 - \tau_{base})\cdot(\cos(\pi k/K)+1)/2$ where $k$ is the current iteration, and $K$ is the total number of training iterations. For ImageNet we train the ResNet-50 for 300 epochs with $LR_{base} = 0.3$, batch size of 512, $\tau_{base}=0.99$, and weight decay of $1\cdot10^-6$. For 1000 epochs $LR_{base} = 0.2$, batch size of 4096, $\tau_{base}=0.996$, weight decay of $1.5\cdot10^-6$. For CIFAR, 1000 epochs we increase $LR_{base}=1.0$, $\tau_{base}=0.996$ and increase the batch size to 1024, for 300 epochs we further increase the $LR_{base}=1.5$ and revert $\tau_{base}$ back to $0.99$. STL-10, 1000 epochs $LR_{base}=0.45$, $\tau_{base}=0.996$ and maintain the batch size of 1024. Hyperparameter settings for MHE regularization are given in the following \ref{appendix:lambda}.

\subsection{Dataset Processing}
To most appropriately make direct comparison to BYOL, we share identical dataset processing and augmentation procedures, \cite{grill2020bootstrap}.

\subsection{Dataset Split}
When performing self-supervised pre-training we utilize manually created validation sets to select appropriate hyperparameters, given both the ImageNet ILSVRC-2012 dataset, STL-10 and CIFAR-10/100 datasets do not contain validation splits (we hold out the validation set for use as the test set) we manually generate a subset of the training split for validation. 
\begin{itemize}
\item  ImageNet, we took the last 10009 last images of the official Tensorflow ImageNet split as in \cite{grill2020bootstrap}.
\item  CIFAR-10, we take 500 random samples per class from the train set for validation.
\item  CIFAR-100, we take 50 random samples per class from the train set for validation.
\item  STL-10, we train on the unlabeled set, taking 50 random samples per class from the la belled train set for validation.
\end{itemize}

For all datasets we report the \textit{Top-1} accuracy (\%), which is the proportion of correctly classified examples. For ImageNet we also report the \textit{Top-5} accuracy (\%), the proportion of predictions that are within the top 5 best predictions (the 5 predictions with the highest probabilities).

\subsection{Augmentations}

Augmentation procedure is key to the success of self-supervised learning, therefore to compare our performance against BYOL, we employ the same image augmentations reported in \cite{grill2020bootstrap,chen2020simple}. Undertaken sequentially in the following order:

\begin{itemize}
\item  random cropping: a random patch of the image is selected, with an area uniformly sampled between 8\%and 100\% of that of the original image, and an aspect ratio logarithmically sampled between 3/4 and 4/3. This patch is then resized to the target size of $224 \times 224$ for Imagenet or $32 \times 32$ for CIFAR using bicubic interpolation;
\item  optional left-right flip;
\item  color jittering: the brightness, contrast, saturation and hue of the image are shifted by a uniformly random offset applied on all the pixels of the same image. The order in which these shifts are performed is randomly selected for each patch;
\item grayscale: an optional conversion to grayscale. When applied, output intensity for a pixel (r, g, b)
corresponds to its luma component, computed as $0.2989r + 0.5870g + 0.1140b$;
\item Gaussian blurring: for a $224\times224$ image, a square Gaussian kernel of size $23\times23$ is used, with a standard
deviation uniformly sampled over [0.1, 2.0];
\item solarization: an optional color transformation $x \mapsto x \cdot \mathbf{1}_{\{x<0.5\}}+(1-x)\cdot\textbf{1}_{\{x\geq 0.5\}}$ for pixels with values
in $[0, 1]$.
\end{itemize}

\subsubsection{ImageNet}
For the ImageNet ILSVRC-2012 dataset~\cite{he2016deep} we apply the aforementioned augmentations in the following order: random crop and resize to $224 \times 224$; random horizontal flip; color distortion (random sequence of brightness, contrast, saturation, hue augmentations); random greyscale conversion; Gaussian blur; and color solarization. At evaluation, we simplify augmentations, first applying center crop where in ImageNet images are resized to 256 pixels along the shorter side using bicubic resampling, after which a $224 \times 224$ center crop is applied. Following both the training and evaluation augmentations the transformed images are normalized per color channel by subtracting the average color and dividing by the standard deviation. The average color and standard deviations are computed per dataset.

\subsubsection{CIFAR-10, CIFAR-100, STL-10}
For the CIFAR10 and CIFAR100 datasets, we follow the same procedure as ImageNet resizing to $32 \times 32$, while STL-10 resize to $64 \times 64$. For all three datasets we omit the Gaussian blur and solarization as described in~\cite{chen2020simple}. During evaluation CIFAR images are center cropped at $28 \times 28$ then resized to $32 \times 32$, while STL-10 are center cropped at $56 \times 56$ then resized to $64 \times 64$.

\section{Hyperparameter Analysis: MHE Regularizer} \
\label{appendix:lambda}

We briefly report the linear evaluation results on hyperparameter search for the CIFAR-10 dataset trained for 300 epochs on ResNet-18 encoder and linearly evaluated for 80 epochs freezing the encoder weights. The average of three runs with three seeds are reported as with the all ablation studies. All other hyperparameters remain identical in both of the following experimentation cases.

\subsection{Regularizer Weight \texorpdfstring{$\lambda_{MHE}$}{lg}}

Tab.\ref{tab:weight} reports the linear evaluation results of the BYOL procedure pretrained with differing weighting of the MHE regularization $\lambda_{MHE}$ in the loss. We maintain the same hyperparameter range stated in \cite{liu2018learning} from $10^{-3}$ to $10^2$. We report very little difference in performance over all values for $\lambda_{MHE}$ with $10^{-3}$ diminishing back to BYOL baseline given the small contribution. For all MHE regularization we therefore set $\lambda_{MHE}=10$ given our empirical results.
\begin{table}[ht]
\small
\centering
\begin{tabular}{lr}
\toprule
Weight  & Accuracy (\%)\\
\midrule
0.001 & 90.74 \\
0.01  & 91.12\\
1   & 91.20 \\
10    & \textbf{91.34}\\
100   & 91.14 \\
\bottomrule
\end{tabular}
\normalsize
\caption{Linear Evaluation on CIFAR10 with ResNet-18 Encoder trained for 300 epochs for MHE $s=2$}
\label{tab:weight}
\end{table}

\subsection{Regularizer Power \texorpdfstring{$s$}{lg}}

For powers of $s$, we again observe very little variation in performance between settings, and at worst seeing a reasonable improvement of 0.4\%. For angular MHE we report that $s=a2$ performs substantially better on average. Although more computationally expensive we opt to use the angular-MHE $s=2$ for all experimentation. The outlier performance reported at $s=a1$ was primarily due to one poor result and as such we report the half difference between best and worst.

\begin{table}[ht]
\small
\centering
\begin{tabular}{lrr}
\toprule
Method  & Power & Accuracy (\%)\\
\midrule
BYOL (repro)& - & 90.74 \\
BYOL-MHE    & 0 & 91.12\\
            & 1 & 91.24 \\
            & 2 & \textbf{91.34}\\
BYOL-aMHE   & a0 & 91.14 \\
            & a1 & 90.94 $\pm0.34$\\
            & a2 & \textbf{91.64} \\
\bottomrule
\end{tabular}
\normalsize
\caption{Linear Evaluation on CIFAR10 with ResNet-18 Encoder trained for 300 epochs}
\label{tab:mhe}
\end{table}

\subsection{Batch Size Accuracies}

\begin{table}[ht]
\small
\centering
\begin{tabular}{lrr}
\toprule
Method  & Batch Size & Accuracy (\%)\\
\midrule
BYOL (repro)& 1024 & 90.74 \\
            & 512 & 89.10\\
            & 256 & 87.50\\
            & 128 & 85.00\\
BYOL + $\mathcal{L}_{Uni}$ & 1024 & 91.48 \\
            & 512 & 90.22\\
            & 256 & 86.84 \\
            & 128 & 82.42\\
BYOL-MHE    & 1024 & 91.64\\
            & 512 & 90.02\\
            & 256 & 88.10 \\
            & 128 & 85.76\\
\bottomrule
\end{tabular}
\normalsize
\caption{Corresponding top-1\% CIFAR-10 for different batch size}
\label{tab:batch}
\end{table}

\newpage    
\section{Hyperparameters}
\label{appendix:hyperparams}

\begin{lstlisting}
# Augmentation Probability
--jitter_d=0.5
--jitter_p=0.8                
--blur_sigma=[0.1,2.0]        
--blur_p=0.5                  
--grey_p=0.2                  

# Model
--model=resnet50
--h_units=4096
--o_units=256
--norm_layer=nn.BatchNorm2d
--optimiser=lars
--learning_rate=0.2
--weight_decay=1.5e-6
--batch_size=4096

# Training
# BYOL
--tau=0.996

# MHE Reg
--proj_reg=mhe
--proj_pow=a2
--pred_reg=mhe
--pred_pow=a2
--reg_weight=1

# Fine Tune / Linear Eval 
--ft_epochs=80
--ft_batch_size=100
--ft_learning_rate=0.2
--ft_weight_decay=0.0
--ft_optimiser=sgd

\end{lstlisting}

\end{document}